\documentclass[12pt]{l4dc2021} 

\title[How Are Learned Perception-Based Controllers Impacted by the Limits of Robust Control?]{How Are Learned Perception-Based Controllers Impacted by \\ the Limits of Robust Control?}
\usepackage{times}
\usepackage[font=small,labelfont=bf,figurename=Fig.,tablename=Tab.]{caption}

\usepackage[font=small,labelfont=bf,tableposition=top]{caption}

\DeclareCaptionLabelFormat{andtable}{#1~#2  \&  \tablename~\thetable}

\usepackage{afterpage}

\usepackage{siunitx}

\usepackage{xcolor}

\newcommand{\observation}{z}
\newcommand{\zfreq}{\zeta}

\newcommand{\statedim}{n}

\newcommand{\norm}[1]{\left\|#1\right\|}

\newcommand{\bmtx}{\begin{bmatrix}}
\newcommand{\emtx}{\end{bmatrix}}

 \author{
 \Name{Jingxi Xu} \Email{jxu@cs.columbia.edu}\\
 \addr Columbia University, New York City, NY
 \AND
 \Name{Bruce Lee}\Email{brucele@seas.upenn.edu}\\
 \Name{Nikolai Matni}\Email{nmatni@seas.upenn.edu}\\
  \Name{Dinesh Jayaraman}\Email{dineshj@seas.upenn.edu}\\
\addr University of Pennsylvania, Philadelphia, PA}

\newcount\Comments 
\usepackage{array,multirow,graphicx}
\usepackage{color}
\usepackage{wrapfig}

\definecolor{darkgreen}{rgb}{0,0.5,0}
\definecolor{purple}{rgb}{1,0,1}
\newcommand{\kibitz}[2]{\ifnum\Comments=1\textcolor{#1}{#2}\fi}

\usepackage{booktabs}

\makeatletter
 \let\Ginclude@graphics\@org@Ginclude@graphics 
\makeatother

\begin{document}

\setlength{\belowcaptionskip}{-5pt}


\maketitle

\begin{abstract}%
The difficulty of optimal control problems has classically been characterized in terms of system properties such as minimum eigenvalues of controllability\,/\,observability gramians. We revisit these characterizations in the context of the increasing popularity of data-driven techniques like reinforcement learning (RL), and in control settings where input observations are high-dimensional images and transition dynamics are unknown. Specifically, we ask: to what extent are quantifiable control and perceptual difficulty metrics of a task predictive of the performance and sample complexity of data-driven controllers?
We modulate two different types of partial observability in a cartpole ``stick-balancing'' problem -- (i) the height of one visible fixation point on the cartpole, which can be used to tune fundamental limits of performance achievable by any controller, and by (ii) the level of perception noise in the fixation point position inferred from depth or RGB images of the cartpole.
In these settings, we empirically study two popular families of controllers: RL and system identification-based $H_\infty$ control, using visually estimated system state. Our results show that the fundamental limits of robust control have corresponding implications for the sample-efficiency and performance of learned perception-based controllers. Visit our project website \url{https://jxu.ai/rl-vs-control-web} for more information.
\end{abstract}


\section{Introduction}


Data-driven techniques for robotic control such as deep reinforcement learning have recently become increasingly popular, especially for settings where input observations are high-dimensional, such as images, and state transition dynamics are not known in advance. These techniques have shown great promise for controlling a variety of robots ranging from manipulators~\citep{OpenAI2018-qx} to legged robots~\citep{Lee2020-xx}, drones~\citep{Molchanov2019-gz}, and autonomous cars \citep{kendall2019learning}.  However, these techniques have largely been studied and developed within the confines of 
stylized, often simulated settings, where performance metrics are naturally divorced from important real-world concerns such as safety and robustness.
In the light of recent catastrophic failures of learning-based control systems such as fatal autonomous car collisions~\citep{wp_av-fatality}, we argue that it is imperative to study and characterize the \emph{limitations} of these approaches in challenging settings that present realistic difficulties for observation and control. In particular, how do such difficulties affect the performance and sample complexity of learned controllers?

Classical robust control theory provides a rich set of tools characterizing fundamental limits on achievable performance in terms of system properties such as open-loop unstable poles and zeros. However, analogous theoretical results in the learning-based control literature are not nearly as well developed, especially in the context of controllers that involve high-capacity functional approximators such as deep reinforcement learning from pixels.
Rather than seeking theoretical limits, we try a different tack, empirically studying various families of learned controllers in a setting where control and observation difficulty can be carefully tuned. 




Our empirical study focuses on the visuomotor control task of stabilizing a cartpole in the upright position using only the visual observations from a camera with a head-on view. All visuomotor controllers must implicitly or explicitly solve two important and closely intertwined problems. The first is visual perception, i.e., how to map raw high-dimensional visual observations $o$ to their task-relevant latent causes, denoted as the state representation $x$? The second is the task of synthesizing optimal action policies $\pi(u|x)$ conditioned on those state estimates.
    
In real-world settings, perception is often an underconstrained problem. For example, an autonomous car with on-board cameras cannot see around the corner of a street on a left turn, or a pedestrian occluded behind a parked vehicle, or a pedestrian with dark clothes on a poorly lit street. In all such instances, the observations $o$ are a non-invertible function of the relevant state $x$, and the estimated states $\hat{x}$ output from perception cannot match the state $x$ perfectly, even with the most optimal perception system. This imperfect perception problem may be represented formally as a partially observable Markov decision process (POMDP). The successes of reinforcement learning in the last few years have been largely demonstrated on fully observed tasks, and general methods for tackling POMDPs remain elusive. Even when they are evaluated on real-world robotic systems, robots and environments are typically instrumented to ensure near-complete observability of all relevant state information, which is impractical for in-the-wild applications like autonomous driving.

Our visual cartpole balancing task permits the modulation of two realistic sources of partial observability within the context of a well-studied classical control problem. First, a set fraction of the cartpole's length is constantly occluded from the camera. As we will explain, this type of information loss has been shown to induce fundamental limits on the performance of any controller for this system. Second, the sensing abilities of the camera itself may also limit perception. For example, to estimate the distance from the camera of an object in the scene such as the cartpole, a perception system with access to RGB camera observations would be harder to train, and it would produce more noisy estimates than one receiving inputs from a stereo depth camera.

We study the impact of tuning these sources of perception noise on the performance of two families of learned controllers: system identification-based $H_\infty$ control and reinforcement learning, both using visually estimated system state. Our careful empirical studies clearly show that increasing occlusion and deteriorating sensing quality affect both families of controllers in ways that align well with theoretically predicted limits for classical robust controllers. In particular, sample complexity increases and  final task performance decreases, and the effect of sensing noise is exacerbated as more of the cartpole is occluded from view.

\subsection{Related Work}

\paragraph{Fundamental limits of learning-enabled control.} Much of the research in the learning for control community has focused on characterizing \emph{achievable upper bounds} on the performance of learning based control strategies.  While such upper bounds are common in the literature, \emph{lower bounds} are few and far between. For model-free RL methods as applied to the Linear Quadratic Regulator, such lower-bounds can be found in \citep{tu2018gap}, where the authors derive asymptotic lower bounds on the number of samples needed by both the least-squares-temporal-differencing estimator for policy evaluation, and policy gradient methods for policy improvement.  In \citep{simchowitz2018learning, simchowitz2020naive} information-theoretic lower bounds \citep{tsybakov2008introduction,duchi2016lecture} are derived for arbitrary estimators, which in turn are used to show the optimality of certainty equivalent control for the LQ problem.  We note that in both cases, such lower bounds are only available for \emph{full-state} observation settings.  
Most similar in spirit to this paper in \citep{bernat2020driver}, the authors empirically investigate the effects of loss of controllability and increased instability on the sample-complexity of policy gradient and certainty equivalent based methods. In \citep{venkataraman2019recovering}, the authors show that traditional gradient descent methods converge to solutions with poor margins if applied to the counter-example system of \citep{doyle1978guaranteed}: however, they use analytic expressions in lieu of stochastic approximations of the gradients, and thus do not explore questions related to sample-complexity.  To the best of our knowledge, there have been no investigations of the effects of such fundamental limits on the sample-complexity and generalizability of perception-based learning-enabled control methods.


\paragraph{Reinforcement learning under partial observability.} Reinforcement learning (RL)-based approaches are most commonly studied in settings where the full Markov state information is available to the controller. 
When observations are noisy or incomplete, RL settings are typically framed as \emph{partially observed} Markov decision processes (POMDP)~\citep{jaakkola1995reinforcement}. In POMDPs, the system state $x_t$ at time $t$ is no longer available for training and running a standard RL control policy $\pi(u_t|x_t)$. In its place, all we have are observations $o_t$, which are non-invertible functions of the state $x_t$. To adapt standard RL algorithms to work in POMDPs, two broad families of approaches have been studied: memory-based RL and belief state RL. 
In memory-based approaches~\citep{mccallum1993overcoming,Hausknecht2015-di,zhu2017improving}, the input to the controller is no longer just the current observation, but instead the full history of observations and actions, so that the policy function is $\pi(u_t | o_{\leq t}, u_{<t})$. Truly infinite history may become computationally intractable, so that a limited history window of size $H$ may sometimes be used, containing only the last $H$ observations and actions. In belief state RL~\citep{Kaelbling1998-tn,gregor2019shaping,gangwani2020learning,weisz2018sample,Igl2018-sm}, a variable $x_t$ is typically explicitly nominated by the control engineer as the Markov state based on knowledge about the task, and the conditional distribution $p(x_t | o_{\leq t}, u_{<t})$ is maintained and updated with every new observation, as in standard recursive filtering approaches for state estimation. This conditional distribution, called the ``belief state'' $b_t$, is treated as the sufficient statistic of the full history for determining the optimal next action, so that the policy function is $\pi(u_t | b_t)$. This is equivalent to running RL on a newly constructed fully observed Markov decision process (MDP), called the ``belief MDP'', whose states are the beliefs $b_t$.  To the best of our knowledge, RL approaches for POMDPs have not thus far been systematically evaluated under realistic sources of incompleteness or noise in high-dimensional visual observations. In recent works proposing RL algorithms for POMDPs, evaluation is performed exclusively with synthetic noise added to low-dimensional state observations, or with ``flickering'' visual observations of Atari games~\citep{Hausknecht2015-di,zhu2017improving,Igl2018-sm}. These evaluations neither resemble real perceptual difficulties, nor involve controlled experiments where the degree and type of partial observability is tuned. We address these gaps in our work.

\section{Preliminaries}
\label{sec:prelim}
\paragraph{Robust Control and Fundamental Limits.}
\label{sec:rc}
Consider a single-input, single-output (SISO) linear-time-invariant (LTI) system
\begin{align}
    \label{eq:lti-ss}
     x_{t+1} = Ax_t + Bu_t, \quad z_t = Cx_t + Du_t
\end{align}
with transfer function representation given by $z(\zfreq)=C(\zfreq I-A)^{-1}B + D=:P(\zfreq)u(\zfreq)$, obtained via the 
z-transform of \eqref{eq:lti-ss}, and $\zfreq$ is a complex number serving as a frequency parameter.

\begin{wrapfigure}{r}{.33\textwidth}
\centering
\includegraphics[width=.32\textwidth]{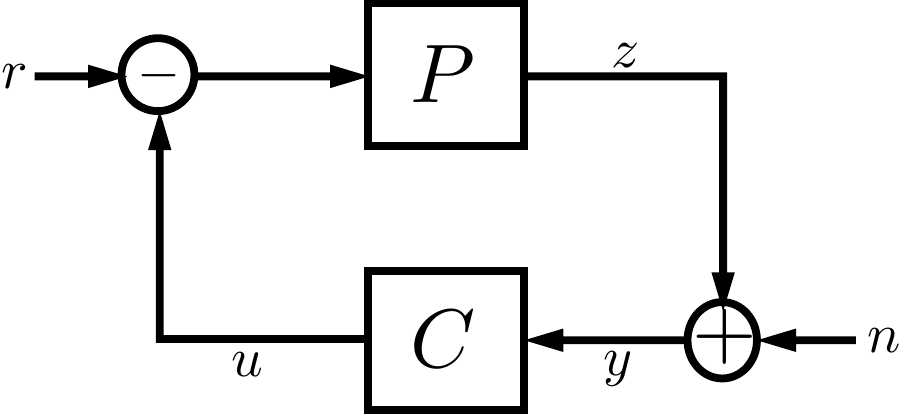}
\caption{Linear feedback control diagram.}
\label{fig:feedback}
\end{wrapfigure}

Furthermore, consider the feedback control system illustrated in Fig.~\ref{fig:feedback}.  In this setup, the reference input $r$ is injected along with the control input $u$ into the system, which produces a controlled output $z$, which the controller $C(\zfreq)$, itself a LTI system, attempts to keep small by using knowledge of the system $P$ and noisy measurements $y = z + n$.  Through straightforward algebra, one can compute that the true output $z$ of the system $P$ under this feedback interconnection is given by
\begin{equation*}
    z(\zfreq) = P(\zfreq)S(\zfreq)r(\zfreq) - T(\zfreq)n(\zfreq), \quad S(\zfreq):= \frac{1}{1 + P(\zfreq)C(\zfreq)}, \quad T(\zfreq) := \frac{P(\zfreq)C(\zfreq)}{1 + P(\zfreq)C(\zfreq)},
\end{equation*}
where $S(\zfreq)$ and $T(\zfreq)$ are the \emph{sensitivity and complementary sensitivity functions} \citep{doyle2013feedback} of the feedback interconnection of Fig.~\ref{fig:feedback}, respectively.  These objects capture the closed-loop maps from reference input $r(\zfreq)$ and sensor noise $n(\zfreq)$ to the regulated output $z(\zfreq)$, and thus through an appropriate quantification of their magnitudes, they can be used as measures of closed-loop performance.  One commonly used measure of system magnitude is the $H_\infty$-norm.
\begin{definition}
The $H_\infty$ norm of a transfer function $G(\zfreq)$ is defined as $ \norm{G}_\infty = \sup_{\omega \in [-\pi, \pi]}|G(e^{j\omega})|.$
\end{definition}
We note that via Parseval's theorem, the $H_\infty$ norm of a system also admits a time-domain interpretation as the worst-case $\ell_2\to\ell_2$ gain of the filter $G_t$ satisfying $\mathcal Z(G_t)=G(\zfreq)$, where $\mathcal Z$ is the z-transform.  As our study focuses on perception-based control, much of our analysis will be focused on characterizing the effects of sensing noise $n(\zfreq)$, and hence, the object of concern will be the $H_\infty$ norm of the complementary sensitivity function $T(\zfreq)$.  To that end, we conclude this section with a useful theorem that allows us to lower bound $\norm{T(\zfreq)}_\infty$ as a function of the open-loop unstable poles and zeros of the system $P(\zfreq)$ for \emph{any possible controller} $C(\zfreq)$, thus establishing limits on achievable performance. A proof may be found in Appendix~\ref{ap:limitsProof}.

\begin{theorem}[Ch.6, \cite{doyle2013feedback}]
Assume that $P(\zfreq)$ has unstable poles $p_k$ and unstable zeros $q_k$, and the interconnection of $P$ with $C$ is internally stable. Then  $\norm{T(\zfreq)}_\infty \geq \max_{i}  \prod_k \left|  \frac{1 - p_i^{-1}q_k^{-1}}{p_i^{-1} - q_k^{-1}}\right|.$
\label{thm:limits}
\end{theorem}
 When $P$ has a single unstable pole and zero, the inequality above simplifies to $\norm{T(\zfreq)}_\infty \geq \left| \frac{pq -1}{p-q} \right|$. Note that as stated, this bound is valid for SISO LTI systems with linear controllers. There are analogous bounds for
 multiple-input, multiple-output (MIMO) LTI systems \citep{goodwin2001control} and LTI systems with nonlinear time-varying controllers  \citep{khargonekar1986uniformly}.

\paragraph{Understanding Limits of Perception-Based Control via Stick Balancing.}
\label{sec:stick}
\begin{figure}[t]
\centering
~~~~~~
\includegraphics[height=100pt]{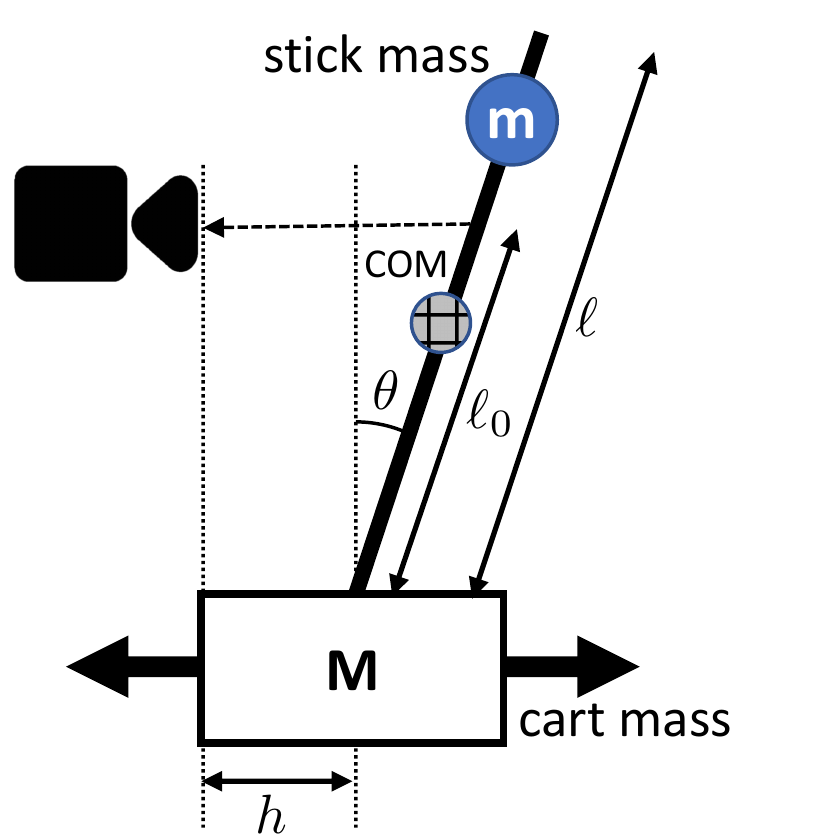}
\hfill
\includegraphics[height=100pt]{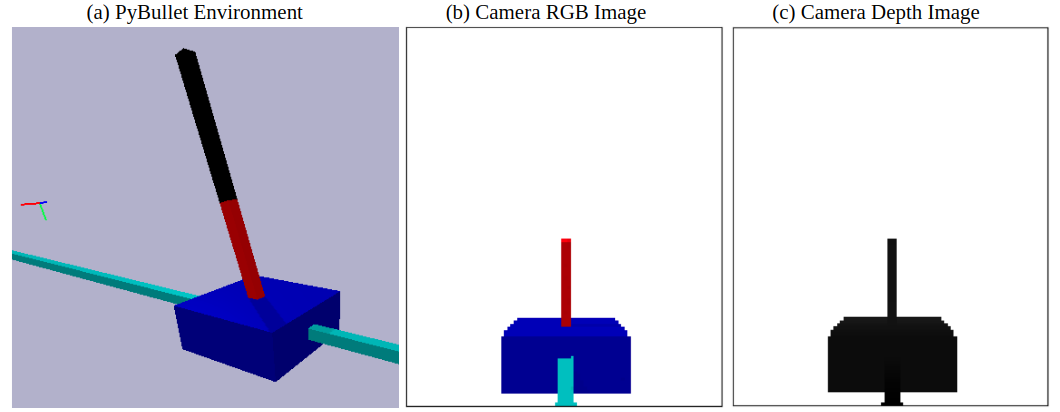}
~~~~~~
\caption{(Left) Schematic of the simplified cartpole system. (Right) Our custom PyBullet environment: (a) a side-on view of the scene to visualize the camera pose with respect to the cartpole system, as well as the occluded portion of the cartpole (in black). (b) RGB image from the camera, (c) depth image from the camera.}
\label{fig:stick}
\label{fig:pybullet}
\end{figure}

 We propose using the ``stick-balancing'' example from \citep{leong2016understanding,doyle2013feedback}, shown in Fig.~\ref{fig:stick} (Left), as a case study.  The dynamics of such a one-dimensional inverted pendulum on a moving cart are given by the following second-order ordinary differential equations:
 \begin{align*}
     (M+m)\ddot{h} + m\ell(\ddot{\theta}-\dot{\theta}^2\sin\theta)=u+r,\, m(\ddot{h}\cos\theta+\ell\ddot{\theta}-g\sin\theta)=0,\, z = h+\ell_0\sin\theta
 \end{align*}
 where $z$ is the distance to point of interest on the pole, which we call the \emph{fixation point}, $u$ is the control force applied to the cart, $r$ is the reference point (set to 0 for all of our experiments), $\theta$ is the pendulum tilt angle from vertical, and $h$ is the horizontal displacement of the cart from the origin.  Additional parameters describing the system are the cart mass $M$, the pole mass $m$, the acceleration due to gravity $g$, the \emph{fixation point} $\ell_0$, and the length of the pole $\ell$.  We further introduce a sensor measurement reading $y= z + n$ for $n$ the sensor noise, to capture the effects of imperfect perception in our model. 
Discretizing the system 
 with Euler integration using a step size of $\tau$ and linearizing the system about the upright position $x=\dot x = \theta = \dot\theta = 0$, one can check that the poles of the resulting linear open-loop system are found at $1$ and $1 \pm \tau \sqrt{\tfrac{(M+m)g}{M\ell}}$.  The system has no zeros if $\ell_0 = \ell$, and zeros at $1 \pm \tau \sqrt{\tfrac{g}{\ell-\ell_0}}$ when $\ell_0 < \ell$. 
 
 This system features two important properties. Firstly, it is unstable, i.e. it has an open-loop unstable pole. When $M\gg m$, this pole satisfies $p\approx 1 + \tfrac{ \tau g^{1/2}}{\ell^{1/2}}$. Second, when $\ell_0<\ell$, it is non-minimum phase, with open-loop unstable zero $q= 1+ \tfrac{\tau g^{1/2}}{(\ell-\ell_0)^{1/2}}$ where we recall that $\ell_0$ denotes the \emph{fixation point} on the stick where the camera is looking (see Fig.~\ref{fig:stick}). It then follows from Theorem \ref{thm:limits} that we can can make the control problem \emph{quantitatively} more difficult by letting $\ell_0\to 0$, i.e., by letting the open-loop unstable zero $q= 1 + \tfrac{\tau g^{1/2}}{(\ell-\ell_0)^{1/2}}$ approach the open-loop unstable pole $p\approx 1+ \tfrac{\tau g^{1/2}}{\ell^{1/2}}$, which in turns implies that $\norm{T(s)}_{\infty} \gtrsim 1/\ell_0.$
We emphasize that for small $|\theta|$, the system is approximately linear so this bound holds \emph{for all possible controllers} \citep{dahleh1994control}, including the optimal $H_\infty$ controller, and thus represents a fundamental lower bound on achievable performance.

\section{Experiments}
In this section, we investigate the performance of data-driven techniques for learning controllers on a customized cartpole balancing environment in PyBullet~\citep{coumans2016pybullet}, which allows for varying fixation lengths $\ell_0$, observation quality, and injecting camera sensing noise. We study two techniques under different measurement models based on learned perception modules: a model-free RL algorithm and a system identification-based robust $H_\infty$ control algorithm. We simulate and vary the fixation point variable discussed above, as well as measurement models and camera sensing noise, in order to tune them to observe their effects on controller performance. Our experiments aim to demonstrate that the performance of learned perception-based controllers is subject to the fundamental limits on achievable performance specified by Theorem \ref{thm:limits}, and answer: What effect does incomplete sensing (as measured by fixation length and the corresponding bounds of Theorem \ref{thm:limits}) and noisy sensing (as measured by the magnitude of simulated camera sensing noise) have on the sample complexity of learning perception-based controllers?




\subsection{Environment}

We developed a custom ``stick-balancing'' environment in PyBullet, illustrated in Fig.~\ref{fig:pybullet}. A cartpole system is actuated by moving the cart along a sliding track (in cyan) along the $x$-axis of the world frame. The mass of the cart $M = 1$\si{kg}, the mass of the pole $m = 0.1$\si{kg}, and the length of the pole $l = 1$\si{m}.

    

The camera has a \emph{head-on} view of the cartpole, as depicted in Fig~\ref{fig:pybullet}(Right, a). To simulate varying fixation points, we occlude a fixed fraction of the pole starting from its top, by setting it to be invisible in PyBullet. This is depicted in black in the simulator view in Fig~\ref{fig:pybullet}(a).  In our experiments, the controller's observations are either direct depth measurements of the fixation point (labeled $z$ in Fig.~\ref{fig:stick}), depth images (Fig~\ref{fig:pybullet}(b)), or RGB images (Fig~\ref{fig:pybullet}(c)). We simulate the parameters of an Intel RealSense D415 camera \footnote{https://www.intelrealsense.com/depth-camera-d415/}, 
and downsample and crop images to $120\times100$ pixels in all our experiments. 

At the beginning of every episode of training and testing, the configuration variables $(x, \dot x, \theta, \dot \theta)$ of the cartpole are all initialized randomly from a uniform distribution over $[-0.05, 0.05]$ (respectively \si{m}, \si{m/s}, \si{rad} and \si{rad/s} for the four variables). This environment is simulated at $50\si{Hz}$ (each time step equals 0.02\si{s}) for 500 steps (10\si{s}) in an OpenAI Gym framework. The episode terminates and resets after 500 steps, or when $x$ goes outside $[-0.6 \si{m}, 0.6 \si{m}]$ or $\theta$ goes outside $[-15^{\circ}, 15^{\circ}]$.

\subsection{Methods}
\label{ss:methods}

\begin{wrapfigure}{r}{.33\textwidth}
\centering
\vspace{-0.2in}
\includegraphics[width=.32\textwidth]{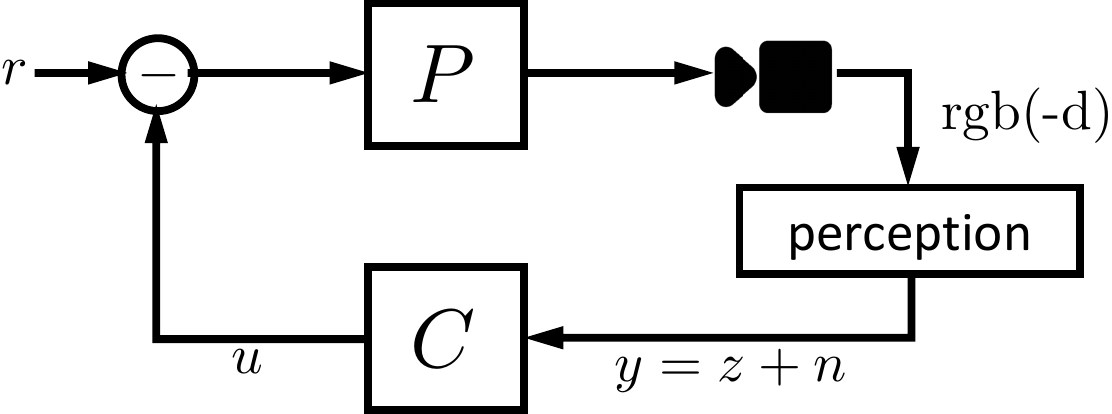}
\caption{Perception-based feedback control diagram.}
\label{fig:visual-feedback}
\vspace{0.05in}
\end{wrapfigure}

\paragraph{Learning Perception Models.} When dealing with images from the camera as input, the role of the perception system is to ``invert'' the observations into an estimate of the depth $z$ of the fixation point, as depicted in Fig~\ref{fig:visual-feedback}. Input images are either depth images or RGB images to allow us to tune partial observability from sensing limitations: we expect that depth images will enable more reliable estimates of $z$ than RGB images.

For each value of the fixation point, and each of depth\,/ \,RGB images, we train a convolutional neural network (see Appendix~\ref{appendix:perception_model} for architecture details) to minimize a mean squared error regression loss on target $z$ labels, using stochastic gradient descent with the Adam optimizer~\citep{kingma2014adam}. Our training set contains 40K images with associated $z$ labels, collected by uniformly sampling $x$ and $\theta$ from the allowable range. Each model is trained up to 1K epochs with early stopping.


As expected, the error in perceiving $z$ from camera images is significantly higher with RGB cameras than with depth cameras. Specifically, the normalized root mean squared errors (RMSE) for estimating the fixation point depth $z$, normalized by the range of $z$, are 0.03\% and 0.25\% for depth and RGB respectively, consistent across different fixation lengths.

\paragraph{Robust Control with System Identification.}

Here we take a classical system identification and robust control approach which consists of first fitting a model to data collected from the system, and then synthesizing an $H_\infty$ controller. 


\begin{wrapfigure}{r}{.4\textwidth}
    \vspace{-0.2in}
    \centering
    \includegraphics[width=.39\textwidth]{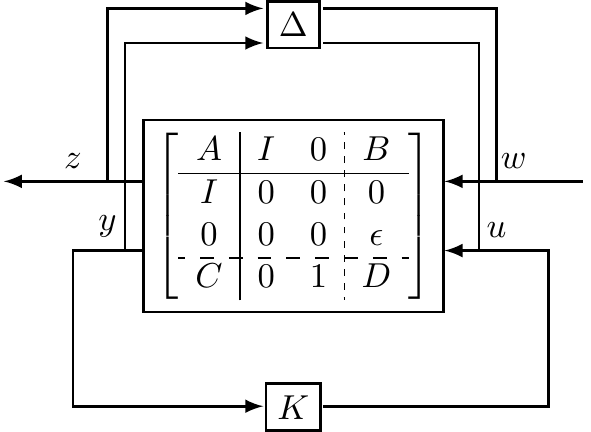}
    \caption{The $H_\infty$ controller $K$ minimizes the worst
    case $H_\infty$ norm of the closed loop system over all uncertainties $||\Delta||_\infty < 1$.}
    \label{fig:mu}
\end{wrapfigure}

\textbf{System Identification:} We first randomly generate system trajectories starting from initial state satisfying $\norm{x(0)}_\infty \leq .05$, and record the resulting depth measurement outputs.  We excite the system with control inputs drawn as $u(t)\overset{\scriptstyle \mathrm{iid}}{\sim{}}U[-10,10]$ until the horizontal position of the cart deviates by more than $0.6$m from its initial position, or the pole deviates more than $15^\circ$ from the vertical.  This data-collection step is repeated for differing numbers of trajectories, for fixation values of $\ell_0 = 1.0, 0.9, 0.8, 0.7$, and for outputs consisting of true depth measurements, depth estimates produced by a perception-map acting on  depth images, and depth estimates produced from a perception-map acting on RGB images.  We also record the full system state, as this will be used to identify a ``baseline'' system model for comparison.  The result is a collection of input/output trajectories $\{z^{(i)}(0:T_i), u^{(i)}(0:T_i)\}_{i=1}^N$ to be used by a system identification algorithm.

We fit a strictly causal 
linear time invariant model \eqref{eq:lti-ss} with parameters $(\hat A,\hat B,\hat C)$ from the collected trajectories $\{z^{(i)}(0:T_i), u^{(i)}(0:T_i)\}_{i=1}^N$.  We apply two system identification methods to the collected data: (i) N4SID, and (ii) a standard two step procedure we refer to as ARXHK. This latter approach consists of first fitting an auto-regressive model $\hat{G}$ of order $p$ by solving the least squares problem
\[
    \sum_{i=1}^N\sum_{t=p}^{T_i}\norm{z^{(i)}(t)- \bmtx z^{(i)}(t-1) &
    & u^{(i)}(t-1) & \dots & z^{(i)}(t-p) & u^{(i)}(t-p) \emtx G}_2^2,
\]
and then applying the Ho-Kalman algorithm~\citep{ho1996effective} to obtain state-space parameters $(\hat{A}, \hat{B}, \hat{C})$. We refer to this latter approach as ARXHK. Exploiting our prior knowledge of the underlying physics of the system, we set the state-dimension (dimension of $\hat A$) to $n=4$. We set the auto-regressive order $p=10$ for ARXHK. 
For the collected data consisting of full state, we fit the state transition matrices $(\hat A, \hat B)$ by solving a least-squares problem
\[
\sum_{i=1}^N\sum_{t=0}^{T_i-1}\norm{x^{(i)}(t+1)-Ax^{(i)}(t)-Bu^{(i)}(t)}_2^2,
\]
and setting $C=[1,\, 0,\, \ell_0, \, 0]$ such that $z = h + \ell_0\theta$. For further details, see Appendix~\ref{ap:sysid}.

\textbf{Robust Control Synthesis:} Once parameters $(\hat A, \hat B, \hat C)$ are identified, we use tools from robust control to synthesize  a controller that can mitigate the effects of uncertainty in the learned model. These uncertainties are caused by noise in the measurements and by approximating nonlinear dynamics with a linear time invariant model \eqref{eq:lti-ss}. To find our controller, we first introduce unstructured uncertainty in feedback with the learned model, as shown in Fig.~\ref{fig:mu}.  We then synthesize an $H_\infty$ controller by drawing on tools from structured singular value, or $\mu$, synthesis~\citep{zhou1996robust} (see Appendix~\ref{ap:controlSyn} for more details).  
The parameter $\varepsilon$ in Fig.~\ref{fig:mu}, which penalizes control effort, is chosen by cross-validation to achieve a suitably high-performing but robust controller. Once a value of $\varepsilon$ has been found to work at the fixation point of $1.00$ for a perception map it is kept fixed throughout the remaining experiments with the perception map. 

\paragraph{Reinforcement Learning from Estimated State.}
We use Soft Actor-Critic~\citep{haarnoja2018soft} for training our RL agents. SAC is a widely used state-of-the-art model-free RL algorithm that has been demonstrated to work well in continuous control settings. The state of the RL agent is the sequence of estimates of the fixation depth value $z$ from the past $H$ steps. At each time step, a new $z$ estimate output by the perception model is appended to the history buffer and the oldest one is abandoned. We set history size $H=200$ based on validation performance. The reward is structured as a survival reward: the agent earns a unit reward for every timestep survived in the environment without episode termination. Since the maximum length of an episode is $T=500$, the maximum achievable reward is $500$. Each agent is trained up to 10K episodes with early stopping. We report results based on 100 trials.   
See Appendix~\ref{ap:rl} for implementation details.

\paragraph{Performance Metrics.} At each fixation depth, for each type of sensor (noise-free observations of the true $z$, depth image observations, RGB image observations), and for each family of learned controller, we report the average reward earned per episode (same as the survival time), over 100 episodes, and also the success rate, which is the fraction of episodes in which the agent survived successfully up to $T=500$.

\subsection{Results}\label{sec:results}

\begin{wrapfigure}{r}{.5\textwidth}
\centering
\vspace{-0.3in}
\includegraphics[width=0.5\textwidth]{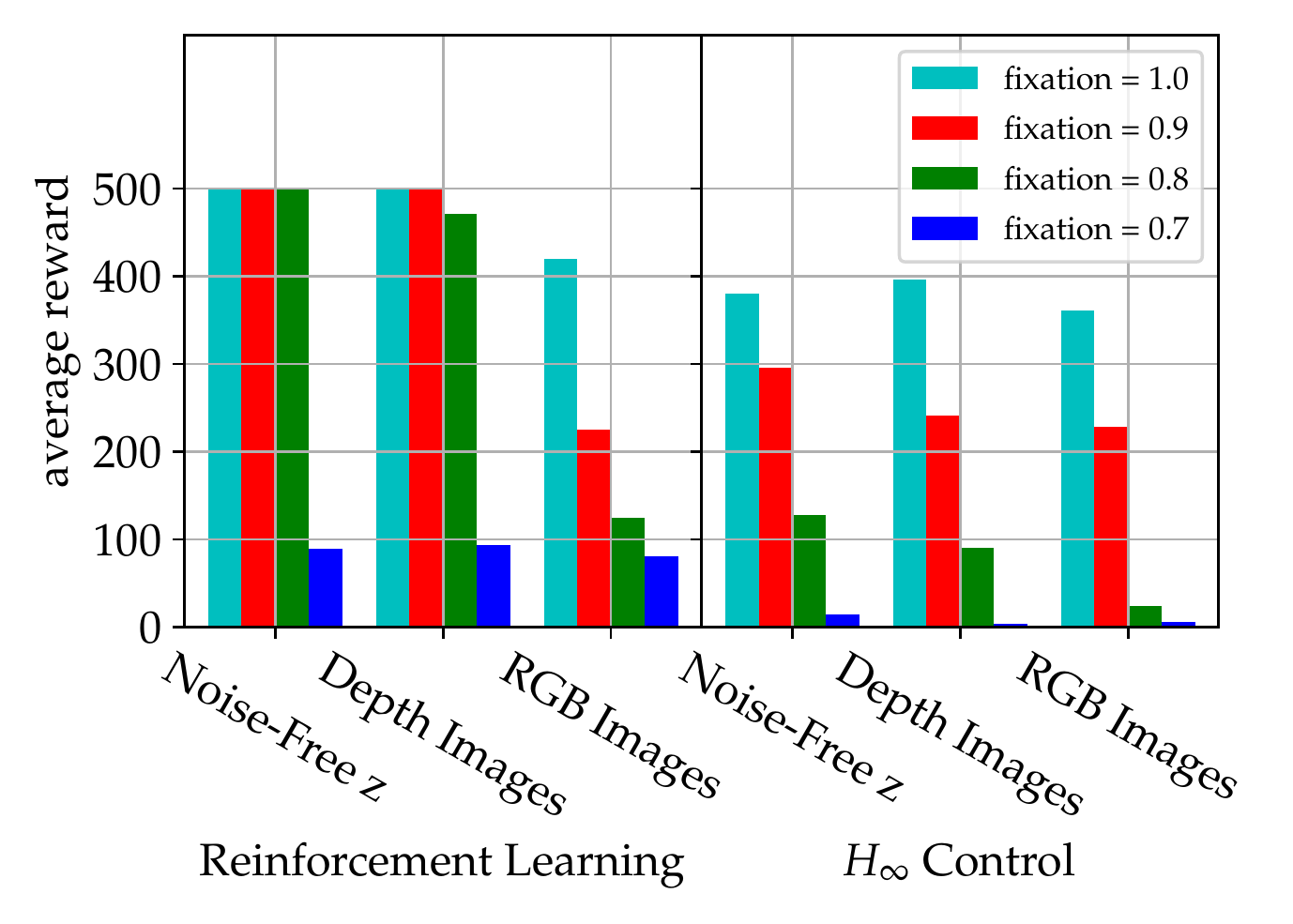}
    \caption{The average reward is plotted for both RL and $H_\infty$ learned perception-based controllers, as a function of fixation height and observation quality. The data is available in Tab.~\ref{tab:main} in Appendix \ref{ap:successdata}.}
    \label{fig:main}
\end{wrapfigure}

Fig.~\ref{fig:main} shows the performance of RL and $H_\infty$ controllers as the fixation height and observation quality are varied. Each RL controller is trained for a maximum of 10K episodes with early convergence, and each $H_\infty$ controller is trained with up to 20K data points used for system identification.

Fig.~\ref{fig:main} presents two main trends. First, performance uniformly deteriorates as the fixation height decreases and more of the pole is occluded from view. This is true for both RL and $H_\infty$ controllers, and at all observation qualities. Next, not only does the performance uniformly worsen as the observation quality deteriorates (from noise-free to depth images to RGB images), but the impact of poorer observation quality is higher at low fixation heights, as can be seen by comparing fixations of 1.0 and 0.8 in the plot. Both these experimental findings are closely aligned with the robustness limitations predicted by Theorem~\ref{thm:limits}, which predicts that sensing noise (which is larger when observation quality is poorer) will be amplified more (as measured by $\norm{T(\zfreq)}_\infty$) at lower fixation points $\ell_0$. 
This suggests that incomplete and noisy sensing act synergistically to further compound the difficulty of the control task when they co-occur.  

\begin{figure}[t]
\includegraphics[width=\textwidth]{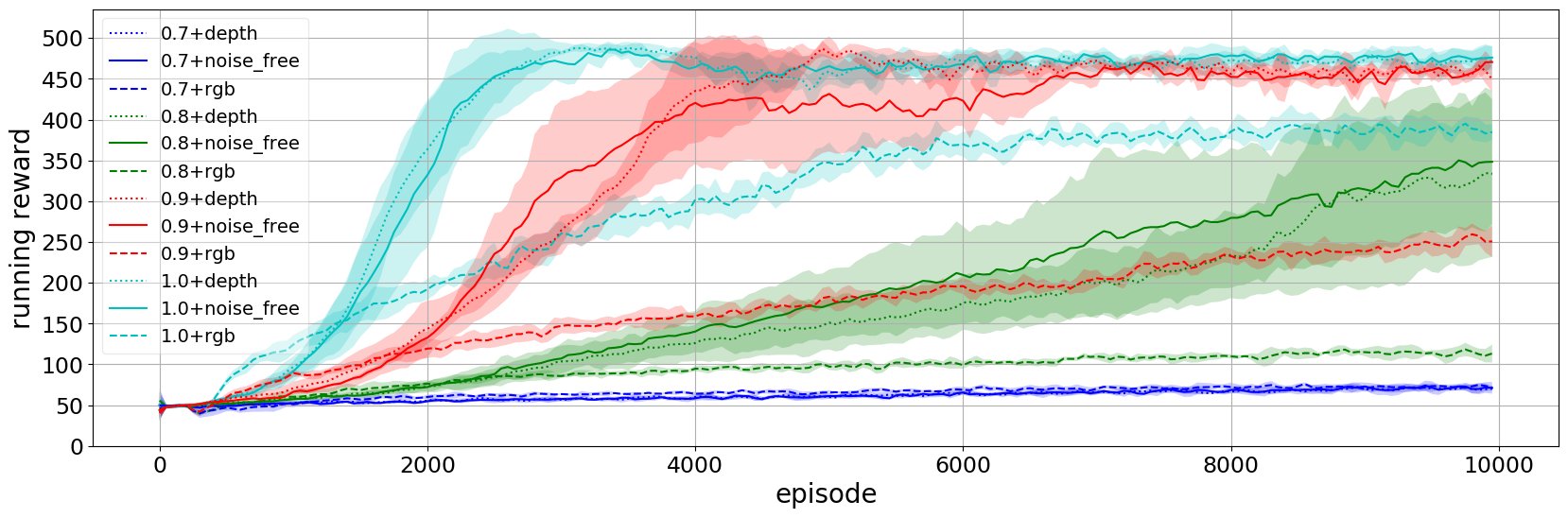}
\caption{Mean and standard deviation of running episode reward from five runs. Plus\,/\,minus one standard deviation is shaded. As the fixation length $\ell_0$ decreases from $1.0$ through $0.7$ and the quality of perception deteriorates from noise-free true fixation point depth to depth images to RGB images, the reinforcement learning agent finds it harder to stabilize the cartpole. 
It takes longer to train, and also achieves lower eventual reward at convergence. 
Furthermore, the effect of perception noise is higher at lower fixation lengths. 
There is less variance in the rewards as the perception noise increases. We conjecture that the small perception noise introduced is acting as a regularizer for training.
}
\label{fig:SAC}
\end{figure}


\begin{wraptable}{r}{.56\textwidth}
  \vspace{-0.2in}
  \small{
  \begin{tabular}{c|c|c|c||c} 
  Model fit & \multicolumn{3}{c||}{\textbf{True state}} & \textbf{Noise-free $z$}
  \\ \hline
    Fixation & Noise-free $z$ & Depth & RGB & Noise-free $z$ \\ \hline
    1.0   & 7.73 & 8.94 & 7.05 & 7.05  \\
    0.9   & 7.33 & 7.28 & 7.16 & 5.84  \\
    0.8   & 7.28 & 6.93 & 6.70 & 5.44  \\
    0.7   & 6.53 & 6.53 & 6.30 & 4.98  
  \end{tabular}}
  \vspace{6pt}
  \caption{Maximum initial angle stabilized by a controller synthesized using a model fit to full states from $100$ trajectories and tested on the three observation scenarios (left) a model fit to true depth measurements from $100$ trajectories using the N4SID algorithm (right).} 
  \label{tab:stabilizedAngles}
  \vspace{-0.2in}
\end{wraptable}

Overall, Fig.~\ref{fig:main} establishes that fixation height and observation quality are very effective at modulating the achievable performance levels for both families of learned controllers. Next, we ask: do these factors also predict the learning speeds for these controllers? Fig.~\ref{fig:SAC} shows the training plots (running average of rewards vs.~the number of training episodes) for the RL agent in each setting. The agent takes longer to learn at lower fixation heights and lower observation quality, and in keeping with the results in Fig.~\ref{fig:main}, it also eventually converges to worse performance.  We conjecture that the increased difficulty in the underlying control problem leads to systems for which only near-optimal policies provide meaningful reward signals, which manifests itself in the increased learning times observed in Fig.~\ref{fig:SAC}: we leave a formal investigation of this phenomenon to future work.

Finally, we investigate why the performance of the $H_\infty$ controller almost uniformly lags behind that of the RL agent in Fig.~\ref{fig:main}. A clue lies in the performance of $H_\infty$ controllers at fixation 0.9, with depth images. We see an average reward of 228.63, with a success rate of 0.31 (see Tab.~\ref{tab:main}). This happens because $H_\infty$ controllers tend to perfectly stabilize the cartpole when it is initialized with small deviations from the vertical, but they fail almost immediately outside this basin of attraction.
To illustrate, the 50th and 75th percentile of $H_\infty$ controller rewards at fixation point $0.9$ using depth images are $3$, and $500$ respectively.

To better understand this phenomenon, we estimate
the maximum initial angles (in degrees) for which 
the $H_\infty$ controllers stabilize the systems.
First, Tab.~\ref{tab:stabilizedAngles} considers
the maximum angle for which a controller 
fit to the \emph{full state} or the true 
z observations successfully stabilizes our system. As these models use full state observations, these quantities serve as a rough upper bound for the maximum initial angles stabilized by controllers synthesized from only noisy observations. 

\begin{figure}
    \centering
     \includegraphics[width = 0.28 \textwidth]{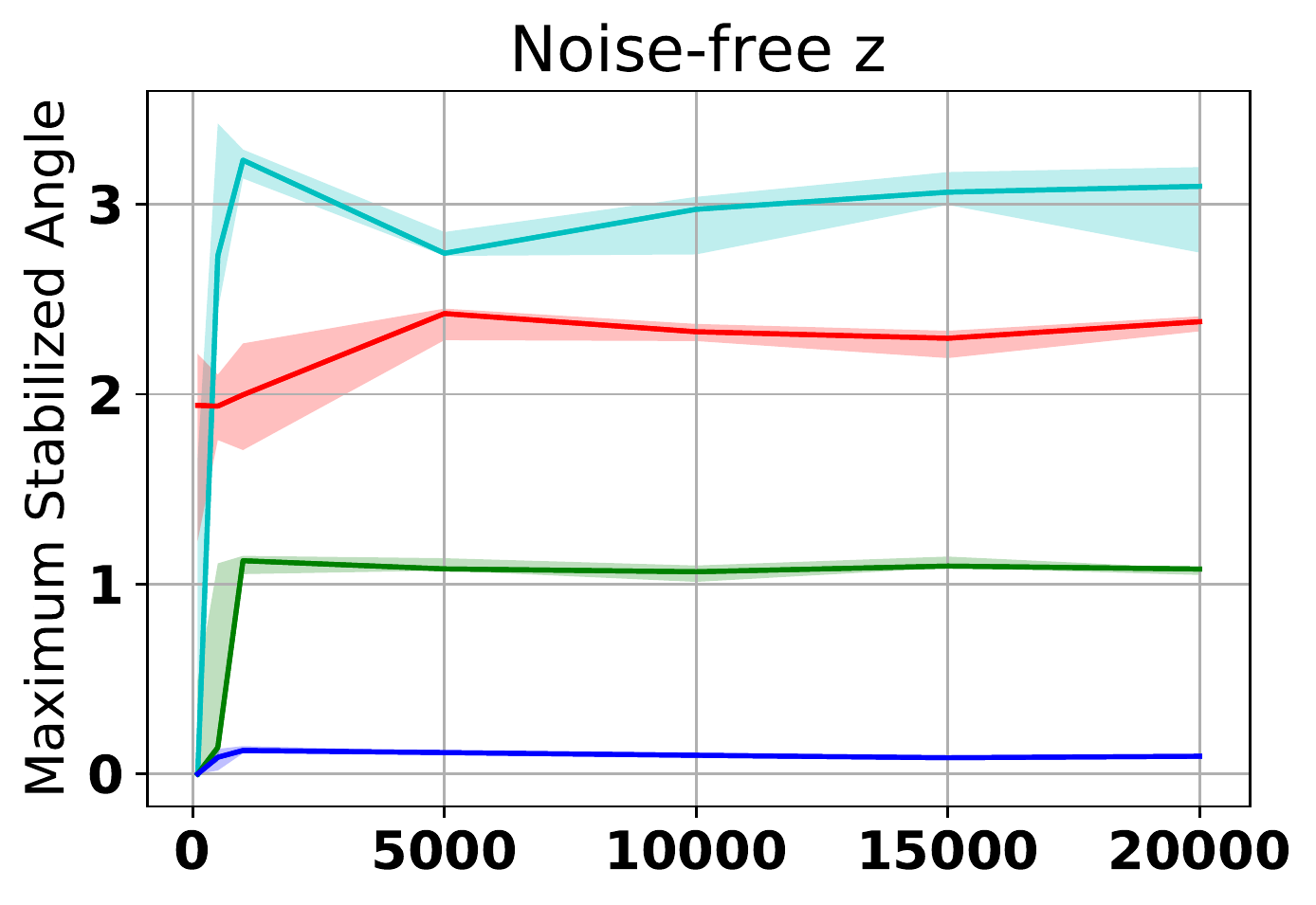}
     \includegraphics[width = 0.28 \textwidth]{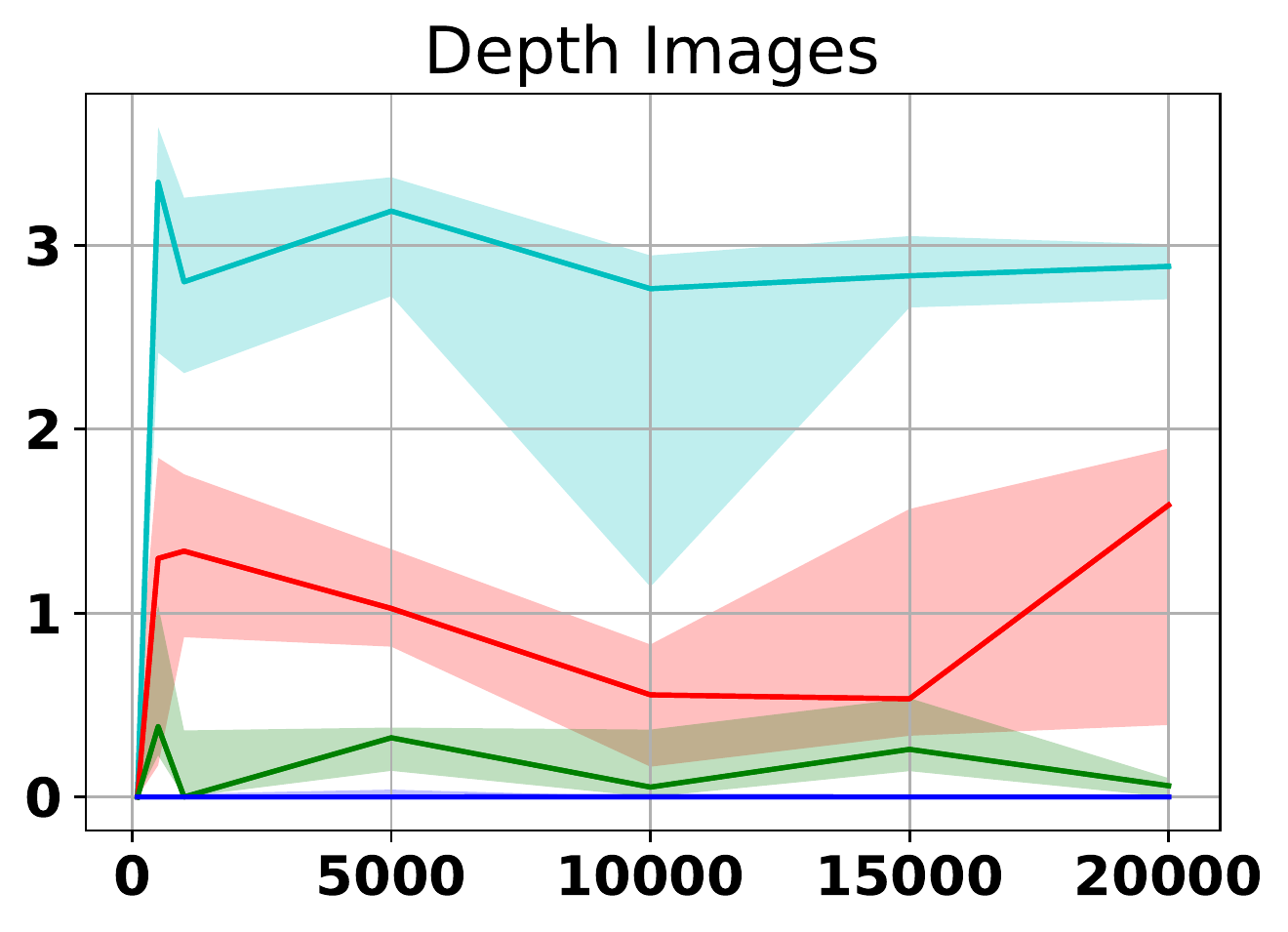}
     \includegraphics[width = 0.28 \textwidth]{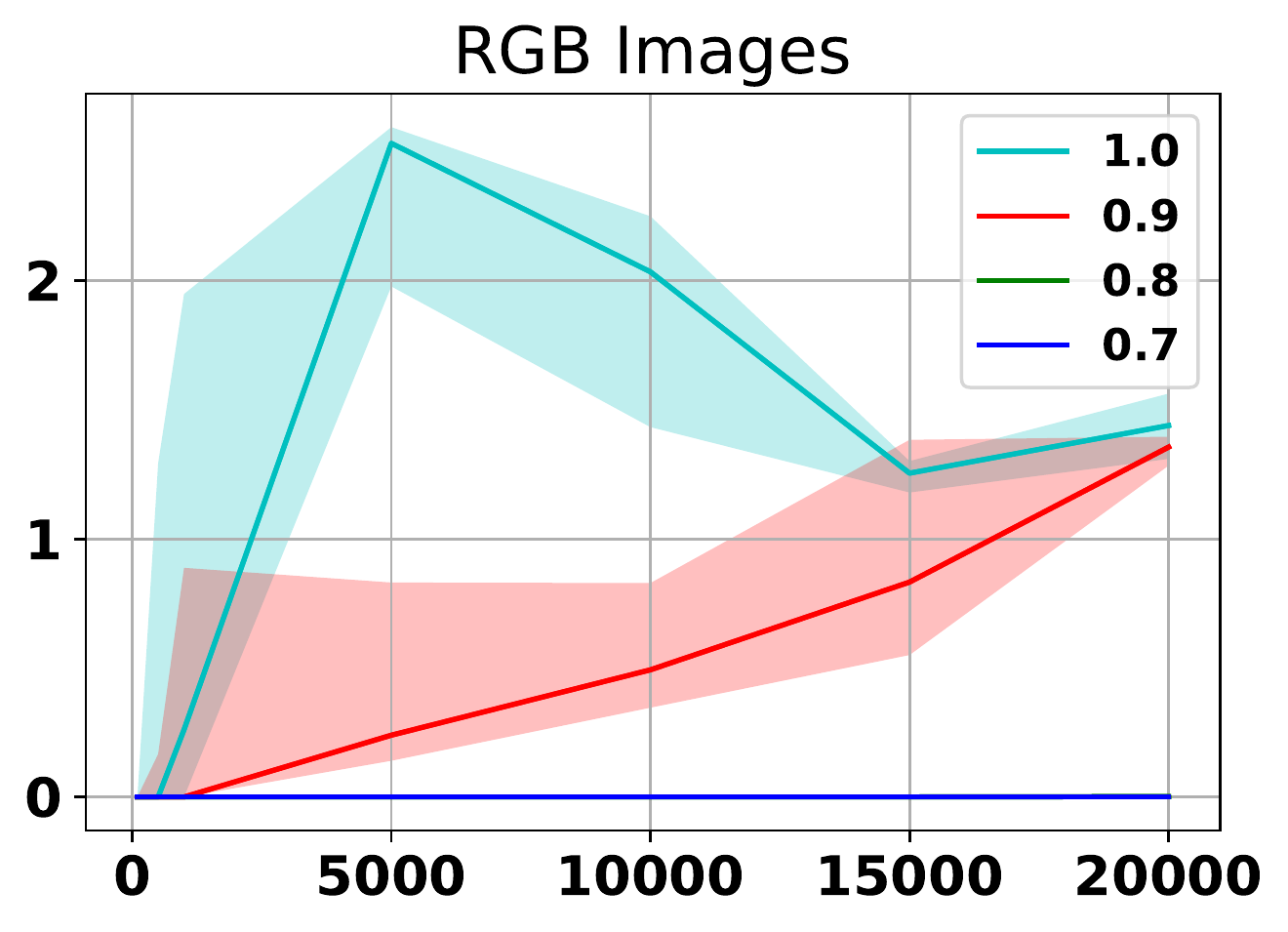}
    \caption{The maximum angles stabilized by the $H_\infty$ 
    controller fit to a model using ARXHK with varying amounts of data. In contrast to Fig.~\ref{fig:SAC}, the x axis here is the number of data samples used by the identification algorithm. For more details on this difference, see Appendix~\ref{ap:sample}. Each curve corresponds to a different fixation point. For RGB images, the green and blue curves overlap.The shaded regions represent the 25th and 75th quartiles across seven random datasets.  }
    \label{fig:controlResults}
    \vspace{-0.1in}
\end{figure}

Now, we measure the stabilizing range of our $H_\infty$ controllers synthesized using ARXHK and noisily perceived $z$ from depth and RGB images.  
For each type of observation, we plot the maximum stabilized angle vs.~the amount of data used to fit the model in Fig.~\ref{fig:controlResults}. 
As the perception problem becomes more difficult
with lower fixation points and noisier $\observation$
measurements, the maximum stabilized angle becomes smaller.  Also of note is the step-like response in the sample-complexity curves of Fig.~\ref{fig:controlResults}: the ARXHK and $H_\infty$ based method required only a few hundred data points to saturate the performance achievable by their model class.  This further suggests that by fitting a slightly richer model (e.g., piecewise linear) and relying on a slightly more sophisticated robust control method (e.g., gain scheduling), the regions of attractions could be expanded to match those of the RL controllers while still requiring much less data.

\section{Discussion}
Through the use of a theoretical model of a simple cartpole system and corresponding customized experimental environment, we have empirically evaluated the consequences of well understood fundamental limits of control on the performance achievable by learned perception-based controllers.  In particular, we examined the effects of limits imposed by unstable dynamics combined with realistic sources of partial observability through incomplete or noisy visual perception.  Our results suggest that these fundamental limits propagate through to other aspects of the learning and control pipeline.  For example, Fig.~\ref{fig:SAC} suggests that training time required to achieve a given level of performance is negatively affected by both poor (low fixation point $\ell_0$) and noisy sensing.  We also observed similar trends in performance (see Tab.~\ref{tab:main}), as measured by reward and success rate for the RL controller, and success rate for the robust $H_\infty$ controller.  We believe that our results are not the consequences of phenomenological behavior unique to the simple system studied in this paper, but that they rather hint at a deeper, more fundamental interplay between how difficult it is to sense and control a system, and how difficult it is to learn to control it. 


\acks{
We thank Natalie Bernat and John C. Doyle for helpful feedback and comments.
}
\bibliography{refs}

\appendix


\section{Proof of Theorem~\ref{thm:limits}}
\label{ap:limitsProof}
The continuous-time analog of this result can be found in Chapter 6 of \cite{doyle2013feedback}.
\begin{proof}
    Assume no unstable pole/zero cancellations between the plant $P(\zfreq)$ and the controller $C(\zfreq)$.  Then note that for any unstable pole $p$ of $P(\zfreq)$, 
    \[
        S(p) = \frac{1}{1+P(p) C(p)} = \frac{1}{\infty}=0,  \, T(p) = 1 - S(p) = 1 
    \]

    We say that a transfer function is \textit{all-pass} if it is one on the unit disc. 
    Any all pass function can be expressed, up to a sign, as a product of 
    factors of the form
    \[
        \frac{\zfreq^{-1} - \bar{a}}{1 - \zfreq^{-1} a}, \, |a| < 1
    \]
    A transfer function is said to be \textit{minimum phase} if it has no zeros outside of the unit disc. 
    We see that any transfer function $G$ may be factored as $G = G_{mp}G_{ap}$ where
    $G_{ap}$ is all-pass and $G_{mp}$ is minimum phase. In particular,
    $G_{ap}$ will receive the unstable zeros of $G$.
    
    Using the fact that for any unstable pole $p_i$, $T(p_i) = 1$, we have
    \[ 
        T_{mp}(p_i) = T_{ap}^{-1}(p_i) = \prod_k \frac{1 - p_i^{-1}q_k^{-1}}{p_i^{-1} - q_k^{-1}}= \prod_k \frac{1 - p_i^{-1}q_k^{-1}}{p_i^{-1} - q_k^{-1}}
   \]
    
    Then by the maximum modulus theorem
    \[
        \|T\|_\infty = \sup_{\omega \in [-\pi,\pi]]} |T(e^{jw})|
        = \sup_{|\zfreq| > 1} |T(\zfreq)| = \sup_{|\zfreq| > 1} |T_{mp}(\zfreq)| = \|T_{mp}(\zfreq)\|_\infty \geq \prod_k \left| \frac{1 - p_i^{-1}q_k^{-1}}{p_i^{-1} - q_k^{-1}} \right|
    \]
    Taking the maximum over the unstable poles provides the desired inequality.
    
\end{proof}
\section{System Identification Methods}
\label{ap:sysid}
\subsection{ARXHK}


Given the collection of input output trajectories 
$\{z^{(i)}(0:T_i), u^{(i)}(0:T_i)\}_{i=1}^N$, an autoregressive order $p$, and model order $\statedim$ , we will 
estimate a linear model $(\hat{A}, \hat{B}, \hat{C})$.
As an intermediary step, we identify the observer $(\hat{A} - \hat{L} \hat{C}, \bmtx \hat{B} & \hat{L} \emtx , \hat{C})$ under
the assumption that $\hat{A} - \hat{L} \hat{C}$ is stable. To simplify notation, define $\tilde{A} :=
\hat{A} - \hat{L} \hat{C}$. In our experiements, ARXHK was
used with $p = 10$, $\statedim = 4$.

First fit an autoregressive model $\hat{G}$ as 
\[
    \hat{G} = 
    \min_G \sum_{i=1}^N\sum_{t=p}^{T_i}\|{z^{(i)}(t)- \bmtx z^{(i)}(t-1) &
    & u^{(i)}(t-1) & \dots & z^{(i)}(t-p) & u^{(i)}(t-p) \emtx G}\|_2^2,
\]
Next define the Toeplitz matrix 
\begin{align*}
    \mathcal{H} = \bmtx \hat{G}(2p-1) & \hat{G}(2p) & \hat{G}(2p-3) & \hat{G}(2p-2) & \dots & \hat{G}(1) &  \hat{G}(2) \\
                        0 & 0 & \hat{G}(2p-1) & \hat{G}(2p) & \dots & \hat{G}(3) &  \hat{G}(4) \\
                        \vdots & & & \ddots & & & \\
                 0 &  & \dots & & 0  & \hat{G}(2p-1) & \hat{G}(2p) \emtx 
\end{align*}

Identify the elements of $\hat{G}$ with the Markov parameters of our desired observer and use the fact that $\tilde{A}$ is assumed
to be stable so that $\hat{C} \tilde{A}^p \hat{B} \approx 0$ for $p$ sufficiently large. Then
\begin{align*}
    \mathcal{H} &=: \bmtx \hat{C} \tilde{A}^{p-1} \hat{B} & \hat{C} \tilde{A}^{p-1} \hat{L} & \hat{C} \tilde{A}^{p-2} \hat{B} & \hat{C} \tilde{A}^{p-2} \hat{L} &\dots & \hat{C} \hat{B} & \hat{C} \hat{L} \\
    0 & 0 & \hat{C} \tilde{A}^{p-1} \hat{B} & \hat{C} \tilde{A}^{p-1} \hat{L} & \dots & \hat{C} \tilde{A} \hat{B} & \hat{C} \tilde{A} \hat{L} \\
    \vdots & & \ddots \\
    0 & & \dots & & 0& \hat{C} \tilde{A}^{p-1} \hat{B} & \hat{C}  \tilde{A}^{p-1} \hat{L}
    \emtx \\
    &\approx 
    \bmtx \hat{C} \tilde{A}^{p-1} \hat{B} & \hat{C} \tilde{A}^{p-1} \hat{L} & \hat{C} \tilde{A}^{p-2} \hat{B} & \hat{C} \tilde{A}^{p-2} \hat{L} &\dots & \dots & \hat{C} \hat{B} & \hat{C} \hat{L} \\
    \hat{C} \tilde{A}^{p} \hat{B} & \hat{C} \tilde{A}^{p} \hat{L} & \hat{C} \tilde{A}^{p-1} \hat{B} & \hat{C} \tilde{A}^{p-1} \hat{L} & \dots & \dots  & \hat{C} \tilde{A} \hat{B} & \hat{C} \tilde{A} \hat{L} \\
    \vdots & & \ddots \\
    \hat{C} \tilde{A}^{2p-2} \hat{B} & \hat{C} \tilde{A}^{2p-2} \hat{L} &  \dots && \hat{C} \tilde{A}^{p} \hat{B} & \hat{C}  \tilde{A}^{p}\hat{L} & \hat{C} \tilde{A}^{p-1} \hat{B} & \hat{C}  \tilde{A}^{p-1} \hat{L}
    \emtx
\end{align*}

If the observability matrix and reversed controllability 
matrix of our desired observer are 
defined as 
\[
    \hat{\mathcal{O}} := \bmtx \hat{C} \\ \hat{C} \tilde{A}  \\
    \vdots \\
    \hat{C} \tilde{A}^{p-1} \emtx, \, \hat{\mathcal{C}} =  \bmtx \tilde{A}^{p-1}\hat{B} & \tilde{A}^{p-1} \hat{L} & \tilde{A}^{p-2} \hat{B} & \tilde{A}^{p-2} \hat{L} & \dots & \hat{B} & \hat{L} \emtx 
\]
respectively, then by the assignment above, $
    \hat{\mathcal{O}} \hat{\mathcal{C}} \approx \mathcal{H}
$.
Then to recover a realization up to some similarity transformation, we may first compute the singular value
decomposition of $\mathcal{H}$ and set
\[
    \mathcal{H} = (U \Sigma^{1/2})(:,1:n) (\Sigma^{1/2} V^\top)(1:n,:) =: \tilde{\mathcal{O}} \tilde{\mathcal{C}}
\]
where MATLAB indexing notation is used in the above expression. 
We may immediately recover estimates for $\hat{B}$, $\hat{C}$ and $\hat{L}$ as $
    \hat{C}:= \tilde{\mathcal{O}}(1:1, :), \, \hat{B}:= \tilde{\mathcal{C}}(:,2p-1:2p-1),\, \hat{L}:= \tilde{\mathcal{C}}(:,2p:2p) $, and solve for 
    $\tilde{A}$ as $ \tilde{A} = \min_{W} \tilde{\mathcal{O}}(:-2) W = \tilde{\mathcal{O}}(2:)$. 
    From $\tilde{A}$ we determine $\hat{A}$ as $\hat{A} = 
    \tilde{A} + \hat{L} \hat{C}$.

\subsection{N4SID}
The tests using N4SID leveraged the 
n4sid function from MATLAB. It was
used with N4Weight set to CVA and N4Horizon
set to [1 10 10]. 
Further details about the algorithm can
be found in~\citep{vanovershee1994n4sid}.

\section{Control Synthesis}
\label{ap:controlSyn}
To account for the uncertainties of the identified model $(\hat{A}, \hat{B}, \hat{C})$ in controller synthesis. 
we consider a disturbance $w_x$ upon the state and $w_y$ upon
the output 
of the system (stacked as $w = \bmtx w_x^\top & w_y^\top \emtx^\top $) and define an auxiliary signal $z$ which 
contains the values we desire to be becomes small, in particular 
the state variables and the actuation effort. The penalty upon
the actuation effort is scaled by $\varepsilon$, a
parameter which can be tuned to achieve a controller with good performance. Accounting for these disturbances in our model, 
and the auxiliary output signal in our model 
results in the following system
\begin{align*}
    x_{k+1} &= \hat{A} x_k + \hat{B} u_k + w_{xk} \\
    y_k &= \hat{C} x_k  + w_{yk}\\
    z_k &= \bmtx I \\ 0 \emtx x_k + \bmtx 0 \\ \epsilon \emtx u_k
\end{align*}
To further enhance the controller's robustness to model 
uncertainties,
we place the resulting system in feedback with 
a unstructured linear time invariant uncertain system. 
This may be performed in MATLAB by calling the feedback function
along with ultisys. 
The controller is then synthesized using the MATLAB function musyn with nmeas and ncont both set to one.


\section{RL and Control Average reward and Success Rate data}
\label{ap:successdata}

\begin{table}[t]
  \small
  \centering
  \vspace{0.06in}
  \resizebox{\textwidth}{!}{
  \begin{tabular}{c|cc|cc|cc}
    \toprule
    Fixation & \multicolumn{2}{c|}{Noise-Free $z$} & \multicolumn{2}{c|}{Depth Images} & \multicolumn{2}{c}{RGB Images}           \\
     & RL & $H_\infty$ & RL & $H_\infty$ & RL & $H_\infty$\\
    \midrule
    1.0   & 500.00, 1.00 & 380.44, 0.76 & 500.00, 1.00 & 396.42, 0.66 & 419.76, 0.58  & 360.80, 0.35                \\
    0.9   & 500.00, 1.00 & 295.56, 0.57 & 499.75, 0.99 & 240.74, 0.41 & 252.05, 0.01  & 228.63, 0.31              \\
    0.8   & 500.00, 1.00 & 127.31, 0.26 & 471.12, 0.69 & 90.17, 0.00 & 124.75, 0.00 & 23.89 ,0.00               \\   
    0.7   &88.79, 0.00   & 13.99, 0.00  & 93.17, 0.00   & 3.80, 0.00  & 80.79, 0.00 & 5.9, 0.00    
    \\\bottomrule
  \end{tabular}
  }
  \vspace{6pt}
  \caption{Learned perception-based controller performance reported as ``average reward, success rate'' for both RL and $H_\infty$ controllers, as a function of fixation height and observation quality.}
  \label{tab:main}
\end{table}

Tab.~\ref{tab:main} has the data from Fig.~\ref{fig:main}.

\section{Controller Synthesis Experimental Procedure}
To obtain the control results in .~\ref{tab:main} and Fig.~ \ref{fig:controlResults} we collected $3000$ runs
with the initial states and actuation signals described in 
Subsection \ref{ss:methods}. For each of the three
perception variations we first determined a value of 
$\varepsilon$ to use in the control synthesis step. To 
do so, we fit a model using ARKHK with $20000$ data points
from a the trials with a fixation point of 1.0 for each perception type,
and varied $\varepsilon$ until the control performance was 
acceptable. These values were fixed for perception
map for the remaining experiments. 
The values of $\varepsilon$ use $5 \times 10^{-3}, 1 \times 10^{-6}$ and $1 \times 10^{-6}$ for the no noise setting,
depth perception map, and the rgb perception map respectively.

Now, for each perception map and each fixation point, 
we fit a linear model using ARXHK auto-regressive horizon using $p=10$, and state dimension $\statedim=4$ using $100, 1000, 5000, 10000, 15000$ and $20000$ data points to fit the model. We then
synthesize a controller according to Appendix~\ref{ap:controlSyn}.
The resulting controller was then tested by running 100 trials
of the PyBullet simulator with max length 500. The number of
times the controller stabilized the system for 500 steps
was recorded, and divided by 100 to determine the success rate.
The average reward was also recorded as the average number
of steps for which the controller stabilized each trial.
An estimate for the maximum initial angular
displacement from which the controller was capable of 
stabilizing the system was also determined by setting all initial states to 
zero and bisecting on the initial angle.
These experiments were repeated seven
times. The medians are plotted and the quartiles are shaded. Only 
the medians for each quantity are reported in Tab~\ref{tab:main}.


\section{Soft Actor-Critic Training Details}\label{ap:rl}
\paragraph{Hyperparamters.} For training the SAC agent with depth image observations, we use temperature parameter $\alpha = 0.2$, target smoothing coefficient $\tau = 0.005$, reward discount factor $\gamma = 0.99$ and learning rate $lr = 0.0003$. We decrease the temperature parameter $\alpha$ to 0.01 for RGB observations to decrease the importance of entropy term against reward. We use Adam~\citep{kingma2014adam} for stochastic gradient descent.

\paragraph{Network architectures.} The input to both the Q-network and policy network is the sequence of estimates of the fixation depth value z from the past 200 steps. Both networks are deep neural networks with 2 fully-connected layers, each with 256 hidden units and followed by the ReLU activation. For the Q-network, a third fully-connected layer outputs the Q value. 
For the policy network, a third fully connected layer outputs the action mean and standard deviation. The final action is sampled from the Gaussian distribution defined by this action mean and standard deviation.

\section{Perception Model Architecture}\label{appendix:perception_model}
The perception models are deep convolutional neural networks with the following architecture: $32 \times 3 \times 3$ conv (stride 2) $\rightarrow$ ReLU $\rightarrow$ $64 \times 3 \times 3$ conv (stride 2) $\rightarrow$ ReLU $\rightarrow$ $128 \times 3 \times 3$ conv (stride 2) $\rightarrow$ ReLU $\rightarrow$ $256 \times 3 \times 3$ conv (stride 2) $\rightarrow$ ReLU $\rightarrow$ flatten $\rightarrow$ $1024 \times 64$ fully-connected $\rightarrow$ ReLU $\rightarrow$ $64 \times 1$ fully-connected. For depth observations, the first layer takes 1-channel input, and for RGB, it takes 3-channel inputs.

\section{Data requirements of Control versus RL}\label{ap:sample}

\begin{figure}[t]
\includegraphics[width=\textwidth]{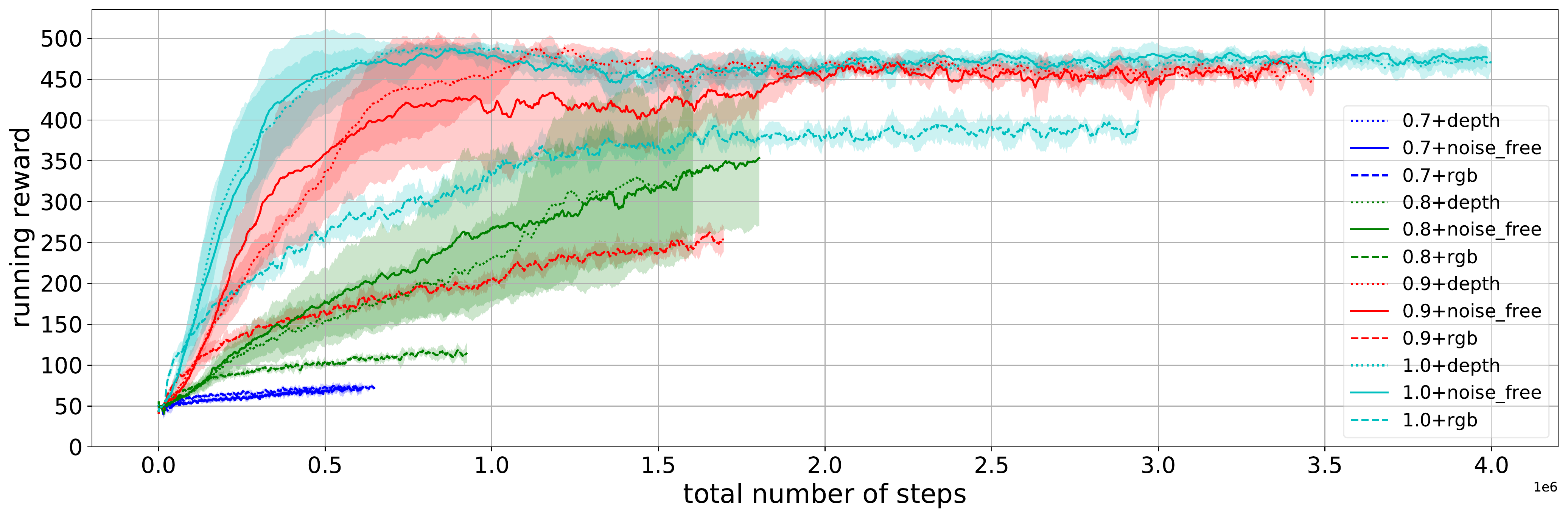} 
\caption{As in Fig~\ref{fig:SAC}, mean and standard deviation of running episode reward from five runs. Plus / minus one standard deviation is shaded. The x axis now is with respect to the total number of steps rather than the number of episodes. The curves for the more difficult environments terminate earlier, as agents that are better able to learn their environment take more steps in a fixed number of episodes.}
\label{fig:cumulative_steps}
\end{figure}

Fig.~\ref{fig:cumulative_steps} is the similar to Fig.~\ref{fig:SAC}, but with the running reward plotted against the total amount of data seen by the RL agent. This allows for a better comparison to the empirical sample complexity of the system identification and control pipeline in Fig.~\ref{fig:controlResults}. In particular, we see that while  system identification followed by control appears to reach its asymptotic behavior in less than 20k data points, each RL agent requires over 500k data points to reach its asymptotic behavior. This is unsurprising, as the current system identification methods fit simple models, resulting in lower data requirements and worse control  performance.

\end{document}